\documentclass[11pt]{article}
\usepackage{coling2018}
\usepackage{times}
\usepackage{url}
\usepackage{latexsym}
\usepackage{graphicx}
\usepackage{multirow}
\usepackage{multicol}
\usepackage{array}
\usepackage{booktabs}
\usepackage[flushleft]{threeparttable}

\title{SHOMA at Parseme Shared Task on Automatic Identification of VMWEs: Neural Multiword Expression Tagging with High Generalisation}

\author{Shiva Taslimipoor and Omid Rohanian \\
  Research Group in Computational Linguistics \\
  University of Wolverhampton, Wolverhampton, UK \\
  {\tt \{shiva.taslimi,omid.rohanian\}@wlv.ac.uk} \\
  }

\date{}

\begin{document}
\maketitle
\begin{abstract}
  
  This paper presents a language-independent deep learning architecture adapted to the task of multiword expression (MWE) identification. We employ a neural architecture comprising of convolutional and recurrent layers with the addition of an optional CRF layer at the top. 
  This system participated in the open track of the Parseme shared task on automatic identification of verbal MWEs due to the use of pre-trained wikipedia word embeddings. It outperformed all participating systems in both open and closed tracks with the overall macro-average MWE-based F1 score of 58.09 averaged among all languages.
  A particular strength of the system is its superior performance on unseen data entries.
\end{abstract}

\section{Introduction}
\label{intro}

Multiword Expressions (MWEs) are linguistic
units consisting of more than one word that are
structurally and semantically idiosyncratic. As
they cross word boundaries \cite{sag2002multiword} and demonstrate multifaceted properties \cite{tsvetkov2014identification}, MWEs don't fit well in the
traditional grammar descriptions where there is a distinct line between lexicon and grammar \cite{green2011multiword,constant2017multiword}.
\newcite{baldwin2005} have reported a notable amount of parsing failures caused by missing MWEs. Simply listing all the potential non-compositional MWEs is not enough due to their complex characteristics including discontinuity, heterogeneity, syntactic variability and most importantly their ambiguity \cite{constant2017multiword}.

MWEs pose greater challenges when processing running texts which has recently attracted more attention \cite{Qu2015,schneider2016semeval}. 
The Parseme shared task on automatic identification of verbal multiword expressions (VMWEs) edition 1.1 \cite{Ramisch-W18} aims at studying automatic labelling of VMWEs in their contexts.  We propose a system that can model the VMWE-annotated training data and predict MWEhood labels for blind test data.  

Previous edition of VMWE shared task \cite{savary2017parseme} features successful systems such as the transition-based \cite{alsaied2017} and the CRF-based systems \cite{maldonado2017}. 
In the transition-based system, sequence labelling is done using a greedy transition-based parser.  
The system, benefiting from dependency tag information of the words, scored the highest in the VMWE shared task edition 1.0 on the blind test data. 
CRF, on the other hand, has become a standard structured prediction method for sequence tagging problems such as NER and MWE identification \cite{vincze2011multiword}.
Structured Perceprtrons have also shown promising results in MWE identification \cite{schneider2014discriminative}. These standard models require handcrafted features to implement. 
Both \newcite{schneider2014discriminative} and \newcite{schneider2016semeval} benefit from the proposed IiOoBb labelling approach for modelling the data.

In recent years, neural network based models, and in particular architectures incorporating Recurrent Neural Networks (RNNs) – such as Long Short Term Memory (LSTM) – and Convolutional Neural Networks (ConvNets) have achieved state-of-the-art performance in sequence tagging tasks  \cite{Collobert2011,lample2016,shao2017character}. \newcite{Gharbieh2017} made the first attempt at using deep learning for MWE identification.
They reported their best results with an architecture consisting of two and three ConvNets. 
The advantages of neural end-to-end sequence taggers 
for MWE identification can be attributed to the use of word embeddings which efficiently encode 
semantic and syntactic information \cite{constant2017multiword}. This is an under-explored but promising area in MWE identification. 

Inspired by the recent synthesis of structured
prediction models and recurrent neural networks
for the task of Named Entity Recognition
(NER) \cite{lample2016,MaHovy2016}, in this work we introduce a similar architecture adapted to the task of MWE identification. 
Specifically, our system benefits from convolutional neural layers as N-gram detectors, bidirectional long-short term memories to take care of long distance relationships between words, and an optional conditional random field (CRF) layer to attend to the dependencies among output tags. 
Our model beats state-of-the-art models across all target languages in terms of generalisability to unseen
MWE types. We achieve this generalisability
without using any task-specific domain knowledge beyond generic POS tags and pretrained embeddings.\footnote{The code is available at \url{https://github.com/shivaat/VMWE-Identification}}


\section{Methodology}

We employ two neural network architectures to tackle the problem of identifying MWEs for multiple languages. The first neural network architecture
that we experiment with is a combination of two ConvNet and one LSTM layers. 
With a view to incorporating ConvNets as n-gram detectors, we apply one convolutional layer with the window size of 2 and the other with size of 3. The features extracted from these two ConvNet layers are concatenated and given to a bi-directional LSTM to model the sequences.

For input representation we benefit from
generic pretrained embeddings (as detailed in Section \ref{sec:wordembeddings}), however in contrast to \newcite{Gharbieh2017}, we combine ConvNets with a bidirectional LSTM. We prevent the model from updating embedding weights in subsequent epochs, and augment the representation of the input with addition of another input layer which receives one-hot representation of POS tags.

Although ConvNets and LSTMs have been
shown to be effective in capturing contextual features and therefore in representing sequences or phrases \cite{Collobert2011}, their simple combinations with final dense layers are blind to the structure of the labels. These models base their tagging decisions on local activation functions like
softmax, and demonstrate lower performance when there are strong interdependencies among output labels \cite{huang2015bidirectional}. 
To effectively predict labels for sequences, a system would benefit from learning the structure of outputs in addition to features of input data. To
this end, we experiment with addition of a CRF layer combined with LSTM and ConvNet. The idea is to use ConvNet and LSTM as feature generation layers of a neural network and CRF as the final output prediction layer. This way we aim to benefit from the individual strengths of all three models in order to effectively tag a corpus for MWEs.

\begin{figure}[!htb]
\centering
\includegraphics[scale=0.3]{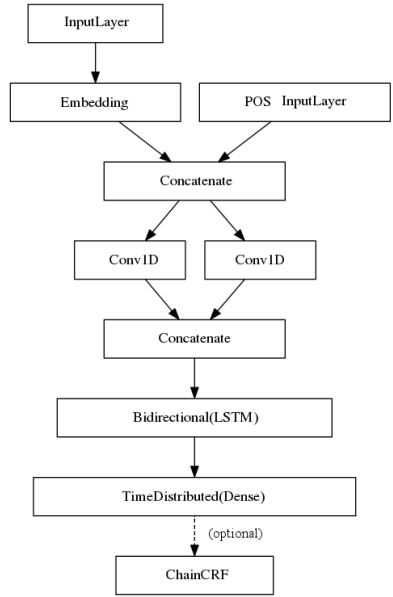} 
\caption{The architecture of ConvNet+LSTM(+CRF) model with two inputs.}
\label{fig:model_outline}
\end{figure}

Addition of a CRF layer on top of a neural network has been experimented with in NER by \newcite{lample2016} and \newcite{MaHovy2016}. The two
studies additionally enhance their network with a character embedding layer. We sidestep this extra layer with the help of pretrained embeddings that take advantage of sub-word information \cite{Bojanowski2016} as explained in Section \ref{sec:wordembeddings}. 
The architecture of our systems is depicted in Figure \ref{fig:model_outline}.

\subsection{Word Embeddings}
\label{sec:wordembeddings}

To learn word representations, an embedding layer can be added to a neural architecture to iteratively learn representation of the data. However, when dealing with small datasets, a better option is to use pretrained embeddings derived from unsupervised methods from larger data. Therefore, in this study we use pretrained embeddings as the input to our neural network architectures and keep them fixed
by preventing the network from updating embedding weights.

Conventional unsupervised methods to construct embeddings from large corpora ignore the internal structure of words and assign a single distinct vector to each word in the corpus.
In the case of rare words in morphologically rich languages, a word with an infrequent inflected form might not receive a generalised representation.
Productivity of these languages result
in many of such low frequency and out-of-vocabulary (OOV) words \cite{sak2010morphology}. On the other hand, even though the inflected form itself
is uncommon, the inflection and other word
formation rules are rule-governed and common.

To alleviate this limitation, \newcite{Bojanowski2016} is an attempt at modelling morphology by integrating subword information. It can be considered
as an extension of the continuous Skipgram
model \cite{mikolov2013distributed}. Inspired by an earlier work by \newcite{schutze1993word}, these word vectors
are learned by representing words as character ngrams and then summing the vectors to derive a full representation for each word. This method is predicated on the hypothesis that character-level
information (including affixes and grammatical rules) contained in the character n-grams help the model develop more generalised semantic representations
and thereby represent rare words better.

For each language, we use a pretrained embedding as described in \newcite{Bojanowski2016} to be used as input to the neural network models. The distributed representations in dimension 300 are obtained with default parameters.
\footnote{The embeddings are available at https:
\url{//github.com/facebookresearch/fastText/blob/master/pretrained-vectors.md}}

\subsection{Features}

We extract additional features in order to complement the input representations. These include 7 binary word shape features. For each token, these features determine whether it starts with a capital
letter, consists of all capital letters, the first character is a \# or @, corresponds to a URL, contains a number or is a digit. The features are added at
the end of embedding vectors for each word.

Furthermore, in order to enrich neural network models with POS information, one-hot representations of POS
tags are given as an additional input (other than word2vec representation) to the neural network models.

\section{Data Prepration}

We focus on the data for the shared task on automatic identification of verbal multiword expressions edition 1.1. It covers data for 20 languages.
Native speakers of different languages annotated both continuous and discontinuous sequences of lexicalised components of VMWEs with different categories of Light verb constructions (LVCs.full and LVC.cause), verbal idioms (VID), inherently reflexive verbs (IRV), verb-particle constructions (VPC.full and VPC.semi), multi-verb constructions (MVC) and inherently adpositional verbs (IAV).\footnote{For more information about categories please refer to \url{http://parsemefr.lif.univ-mrs.fr/parseme-st-guidelines/1.1/?page=categ}. The details of data size are available at \url{http://multiword.sourceforge.net/PHITE.php?sitesig=CONF&page=CONF_04_LAW-MWE-CxG_2018___lb__COLING__rb__&subpage=CONF_40_Shared_Task}}

The PARSEME annotation is composed of the
VMWE's consecutive number in the sentence and its category. The category is marked only for the initial token in a VMWE, for example, 2:VID if a token signals an idiom which is the second VMWE in the current sentence.  
We convert the annotation codes into a labelling format similar to IOB.\footnote{The standard IOB labelling is proposed for chunking where there is no gap between the components of one chunk.
MWE identification task is different since an MWE may or may not be continuous.} To this end, the initial token of a MWE receives the tag $B-$ plus its category.  
Other components of the expression receive the tag $I-$ plus their category. 
Other tokens which are not part of a VMWE -- the ones which are marked with $*$
-- receive the tag $O$. In the datasets, VMWEs rarely overlap. In these cases we follow the same annotation scheme as in the shared task by separating multiple tags for a token with a semicolon.\footnote{Our systems consider this combined tag as a separate category in the list of all possible tags for the data. This introduces
some limitations for the learning system which does not recognise that a candidate MWE can adopt either of these labels individually. On the other hand, this representation can help the system learn what combinations of labels are acceptable.}

In this IOB-like labelling scheme, we differentiate between the beginning component of an expression and its other components. This distinction is not a requirement for the shared task evaluation
(i.e. if any component of an expression is
identified with the correct category it is considered as true positive). However, in the results we show that it is beneficial to have these different tags and perform some filtering based on them. Specifically,
when we convert the labels from prediction
results back to the shared task format, we filter those cases that have label $I-$ without a preceding $B-$ and re-tag them as non-MWEs (by marking them with $*$).

\section{Experiments}
\label{sec:Experiments}

Conforming to one of the main purposes of the shared task which is the development of language-independent and cross-lingual VMWE identification systems, we perform the validation phase and parameter optimisation stage of our systems on \textbf{5} languages, and re-train and test the optimised model for all languages in the shared task.

We evaluate the performance of the system in the validation phase by comparing it to two baseline systems: a standard CRF and \newcite{turian2010}'s CRF which uses word embeddings.

\subsection{Baseline}

Based on \newcite{turian2010}'s CRF approach we first use features that are generally used in CRF implementation as follows:

\begin{itemize}
\item word and lemma form of the current token (in position $i$) and the tokens in the window of size 2 on the left and the right sides of the current word ($w_{i-2}$, $w_{i-1}$, $w_{i}$, $w_{i+1}$, $w_{i+2}$, $l_{i-2}$, $l_{i-1}$, $l_{i}$, $l_{i+1}$ and $l_{i+2}$)
\item POS of the current token and tokens in the window of size 2 on the left and the right sides of the target word ($p_{i-2}$, $p_{i-1}$, $p_{i}$, $p_{i+1}$ and $p_{i+2}$)
\item bigrams of word and lemma forms including tokens in a window of size 1 around the current token ($w_{i-1}w_{i}$, $w_{i}w_{i+1}$, $l_{i-1}l_{i}$, and $l_{i}l_{i+1}$)
\item POS bigrams and trigrams in the window of size 2 around the current word ($p_{i-2}p_{i-1}$, $p_{i-1}p_{i}$, $p_{i}p_{i+1}$, $p_{i+1}p_{i+2}$, $p_{i-2}p_{i-1}p_{i}$, $p_{i-1}p_{i}p_{i+1}$, and $p_{i}p_{i+1}p_{i+2}$)
\end{itemize}

We use Turian's implementation which employs CRFSuite \cite{CRFsuite} with the above features. We also supplement it with pretrained embeddings of words in the widow of size 2 around the current word ($embed_{i-2}$, $embed_{i-1}$, $embed_{i}$, $embed_{i+1}$ and $embed_{i+2}$).
The hyperparameter l2-regularization sigma is set to 2 which is reported to be the optimal value for chunking based on \newcite{turian2010}. 

As a result we have two baselines:
One is the standard CRF and the other is Turian's CRF with word representation features.

\subsection{Neural Network Parameter Settings}
\label{subsec:parameterset}

The details of the layers which are depicted in Figure \ref{fig:model_outline} are presented in this subsection.
These parameters include the number of neurons in each hidden layer, the number of iterations before training is stopped, activation function, and more specifically, the filter size for ConvNet, and the dropout rate for LSTM.

In the first layer every token is represented by its vector from pretrained embeddings concatenated with 7 word shape features.
These are then fed to two ConvNet layers with $200$ neurons, and filter sizes of $2$ and $3$.
These two layers are then concatenated. Since most of the VMWEs are bigram or trigram combinations we find filter sizes of 2 and 3 to be the best choice for extracting n-gram features. However, we also try filter size $5$ for training and see no improvement on the validation set. 
We apply no dropout for the ConvNet layers and use \texttt{relu} as the activation function.
The output of the convolutional layers is given to a bi-directional LSTM with $300$ neurons, dropout of 0.5 and recurrent dropout of 0.2.

In all cases, we get better performance when we set the embedding layer not to be trainable.

\section{Evaluation and Results}

Precision (P), recall (R) and F1-score (F) measures are computed in two settings: one is strict matching (MWE-based) in which all components of an MWE are considered as a unit that should be correctly classified; the other is fuzzy matching (token-based) in which any correctly predicted token of the data is counted \cite{savary2017parseme}.

We compare the results of our models (ConvNet+LSTM and ConvNet+LSTM+CRF) with the baselines (standard CRF and Turian) on the validation data. 
We also make comparisons with the transition-based system \cite{alsaied2017} that scored the highest in the VMWE shared task edition 1.0 using the available
implementation of the model, \footnote{ \url{https://github.com/hazemalsaied/ATILF-LLF.v2}} with the same parameters as they set.
 As annotation schemes differ between versions 1.0 and 1.1, in order to apply the software to the dataset from the newer version of the shared task, the parser disregards MWE types, considering all of them as one type. In this way, we are capable of evaluating the system for overall token-based and MWE-based performance.\footnote{In the first edition of the shared task, in the open track, there was no sufficiently successful system which we can compare our model with. We also aim to show the advantages of using generic extrinsic data.} 
  Furthermore, we report the results of our selected system on the blind test data. We also show the performance of our system considering its important characteristics including its generalisability and the filtering that we use after labelling the data.

\subsection{Validation Results}

\begin{table}[h]
\centering
\small
\caption{\label{t:results-valid} Validation results for the data of shared task edition 1.1.}
\setlength{\tabcolsep}{0.3em}
\begin{tabular}{l l | c c c c | c c c c }
\toprule
\multirow{2}{*}{} & \multirow{2}{*}{Model} & \multicolumn{4}{c}{Token-based} & \multicolumn{4}{c}{MWE-based}  \\
  & & P & R & & F1 & P & R & & F1 \\
\midrule
\multirow{5}{*}{ES} & CRF & 78.62  & 39.35 & & 52.45 & 69.55 & 37.00 & & 48.30 \\
\multirow{4}{*}{} & Turian  & 65.53  & 46.07   &  & 54.10 & 47.11  & 35.80  & & 40.68 \\
\multirow{6}{*}{} & TRANSITION  & 76.02 & 51.81  & & 61.62 & 68.47 & 48.20 & & \textbf{56.57} \\
\multirow{4}{*}{} & ConvNet+LSTM & 71.61 & 58.44 & & \textbf{64.36} & 54.81 & 49.00 & & 51.74 \\		
\multirow{5}{*}{} & ConvNet+LSTM+CRF  & 71.19 & 52.87 & & 60.68 & 58.37 & 47.40 & & {52.32} \\
\midrule
\multirow{4}{*}{EN} & CRF & 58.06  & 12.71 & & 20.86  & 47.67 & 12.39 & & 19.66 \\
\multirow{4}{*}{} & Turian & 59.44  & 15.11 & & 24.10  & 48.98 & 14.50 & & 22.38 \\
\multirow{4}{*}{} & TRANSITION  & 54.32 & 24.86  & & 34.11 & 52.83 & 25.38 & & \textbf{34.29} \\
\multirow{5}{*}{} & ConvNet+LSTM & 55.39 & 31.92 & & \textbf{40.50} & 35.34 & 26.59 & & {30.34} \\
\multirow{5}{*}{} & ConvNet+LSTM+CRF & 52.03 & 27.12 & & 35.65 & 35.59 & 23.87 & & 28.57 \\
\midrule
\multirow{5}{*}{FR} & CRF & 87.02  & 58.85 & & 70.21  & 71.07 & 53.90 & & 61.30 \\
\multirow{5}{*}{} & Turian  & 86.14   & 62.43 & & 72.39  & 71.29 & 57.23 & & 63.49  \\
\multirow{5}{*}{} & TRANSITION & 85.82   & 72.26 & & \textbf{78.46}  & 82.75 & 70.91 & & \textbf{76.37}  \\
\multirow{5}{*}{} & ConvNet+LSTM  & 85.81 & 68.82 & & {76.38} & 77.32 & 66.14 & & {71.29} \\
\multirow{5}{*}{} & ConvNet+LSTM+CRF & 74.69 & 67.13 & & 70.71 & 67.24 & 62 & & 64.52 \\
\midrule
\multirow{5}{*}{DE} & CRF & 82.23  & 28.90  & & 42.77  & 43.69 & 19.36 & & 26.83 \\
\multirow{5}{*}{} & Turian  &  75.85  & 33.53 & & 46.51  & 42.70 & 23.35 & & 30.19  \\
\multirow{5}{*}{} & TRANSITION  &  73.24  & 46.02 & & 56.52  & 68.24 & 46.31 & & \textbf{55.17}  \\
\multirow{5}{*}{} & ConvNet+LSTM & 70.75 & 45.32 & & 55.25 & 47.67 & 38.72 & & 42.73 \\
\multirow{5}{*}{} & ConvNet+LSTM+CRF & 60.27 & 54.08 & & \textbf{57.01} & 41.99 & 45.51 & & {43.68}\\
\midrule
\multirow{5}{*}{FA} & CRF & 94.53  & 77.54  & & 85.20  & 87.07 & 75.25 & & 80.73 \\
\multirow{5}{*}{} & Turian  & 93.88   & 78.74 & & \textbf{85.64}  & 86.65 & 76.45 & & 81.23  \\
\multirow{5}{*}{} & TRANSITION  & 90.84 & 79.10 & & 84.57 & 86.50 & 78.04 & & \textbf{82.06} \\
\multirow{2}{*}{} & ConvNet+LSTM & 91.80 & 78.00 & & 84.34 & 81.74 & 73.25 & & 77.26 \\
\multirow{2}{*}{} & ConvNet+LSTM+CRF & 91.86 & 76.54 & & 83.50 & 81.33 & 73.05 & & 76.97 \\
\bottomrule
\end{tabular}
\end{table}
The evaluation results on the validation sets are presented in Table \ref{t:results-valid}.
We see consistent improvement compared to the baselines (CRF and Turian) in all languages except for Persian (Fa). 
The good performance of the baselines for Persian can be attributed to the fact that in the Persian dataset, MWEs are all LVCs which follow a rigid structure. They are usually constructed from a small number of verbs plus nouns. Also, the proportion of seen MWE types is comparatively high.\footnote{Refer to the details of data in the table in \url{http://multiword.sourceforge.net/PHITE.php?sitesig=CONF&page=CONF_04_LAW-MWE-CxG_2018___lb__COLING__rb__&subpage=CONF_40_Shared_Task}} 
This makes their detection fairly easy for a pattern-based model. 

The performance of our neural network model is better than the TRANSITION system in terms of token-based F-measure for ES, EN and DE. However, the transition-based system works the best in terms of MWE-based for the 6 languages. 
%
%
As for our proposed ConvNet+LSTM+CRF model, we do not see much improvement in general across all languages over the initial ConvNet+LSTM model. 
The significant difference between results of the two systems is mostly in the case of per token evaluation, based on which ConvNet+LSTM usually performs better. In only two out of the five languages, ConvNet+LSTM+CRF shows better results in terms of MWE-based F1 score (DE and ES). In German (DE) we also see a slight improvement in terms of token-based F1 score. In cases where MWE-based score is slightly higher for ConvNet+LSTM+CRF, this small improvement is accompanied by the same or higher amount of drop in token-based F1. In these cases there is a trade-off between token-based and MWE-based evaluation measures.\footnote{We initially planned to include \newcite{lample2016}'s model as an additional baseline. However, applying an implementation of that model to our data resulted in a performance well below other baselines. This suggests that such architectures need to be fine-tuned to our specific task. We also experimented with simpler neural models similar to \newcite{Gharbieh2017}, without observing a significant improvement over our baselines.} 

Based on the overall results obtained in the validation phase, and since we require that the model be applied independently of any particular language, we choose to run the first model, ConvNet+LSTM, on the blind test data for all languages.

\subsection{Test Results}


\begin{table}[h]
\centering
\small
\begin{threeparttable}
\caption{\label{t:results-test} Performance results (F1) on the Test data.}
\setlength{\tabcolsep}{0.4em}
\begin{tabular}{l l c c c c c c c c c c c c c c c c c}
\toprule
 & & & & & & & & &
 \multicolumn{2}{c}{LVC.full} & \multicolumn{2}{c}{VID} & \multicolumn{2}{c}{VPC.full} & 
 \\
\multirow{1}{*}{} & \multirow{1}{*}{} &  \multicolumn{3}{c}{Token-based} & \multicolumn{3}{c}{MWE-based} & \# & T & M & T & M & T & M &    \\
 & & P & R & F1 & P & R & F1 &  \\
\midrule
BG &  & 79.98 & 56.43 & \textbf{66.17} & 75.6 & 55.97 & \textbf{64.32} & 2   & 
50.36 & 49.88 & 40.21 & 34.78 & - & -  \\
\midrule
\multirow{1}{*}{DE*} & & 73.64  & 44.41 & \textbf{55.41} & 54.32 & 40.36 & \textbf{46.31} & * & 15.26 & 6.35 & 48.03 & 36.31 & 68.75 & 59.79 \\
\midrule
EL &  & 78.6 & 68.08 & \textbf{66.79} & 62.79 & 53.89 & \textbf{58} & 1 & 68.56 & 60.03 & 56.1 & 50 & 24.24 & 11.11 \\
\midrule
\multirow{1}{*}{EN*} &  & 60.36 & 18.77 & \textbf{28.63} & 48.40 & 18.16 & \textbf{26.42} & * & 11.23 & 6.25 & 11.76 & 14.74 & 46.67 & 45.09 &  \\
\midrule
\multirow{1}{*}{ES} &  & 38.33 & 53.57 & \textbf{44.69} & 31.65 & 48.8 & \textbf{38.39} &  2 & 30.29 & 22.11 & 32.82 & 30.77 & 0 & 0  \\
\midrule
EU & & 86.27 & 74.95 & \textbf{80.21} & 81.8 & 72.8 & \textbf{77.04} & 1 & 82.16 & 79.13 & 58.78 & 57.81 & - & - \\
\midrule
\multirow{1}{*}{FA} &  & 93.87 & 74.3 & \textbf{82.95} & 86.12 & 71.86 & \textbf{78.35} &	1 & 82.89 & 78.43 & - & - & - & -   \\
\midrule
\multirow{1}{*}{FR*} &  & 82.94 & 57.73 & \textbf{68.08} & 72.39 & 54.22 & \textbf{62} & 1* & 62.82 & 58.09 & 67.84 & 62.47 & - & -\\
\midrule
HE & & 74.43 & 31.26 & \textbf{44.02} & 61.37 & 28.49 & \textbf{38.91} & 1 & 48.03 & 43.4 & 41.61 & 34.38 & 0 & 0 \\
\midrule
HI & & 84.3 & 68.57 & \textbf{75.62} & 76.84 & 69 & \textbf{72.71} & 2 & 76.16 & 71.73 & 10.32 & 12 & - & - \\
\midrule
HR & & 79.52 & 45.88 & \textbf{58.19} & 59.58 & 39.96 & \textbf{47.84} & 2 & 29.97 & 24.34 & 17.48 & 14.63 & - & - \\
\midrule
HU & & 93.49 & 80.87 & \textbf{86.73} & 90.08 & 81.96 & \textbf{85.83} & 3 & 70 & 59.36 & 90.91 & 84.21 & 92.39 & 91.72 \\
\midrule
IT* & & 67.55 & 49.30 & \textbf{57} & 40.09 & 43.55 & \textbf{46.15} & * & 56.28 & 45.60 & 43.59 & 34.03 & 47.37 & 42.11  \\
\midrule
LT & & 56.76 & 18.7 & \textbf{28.13} & 35.74 & 16.8 & \textbf{22.86} & 3 & 39.04 & 29.14 & 9.09 & 6.98 & - & - \\ 
\midrule
PL & & 79.32 & 57.13 & \textbf{66.42} & 73.05 & 56.31 & \textbf{63.6} & 2 & 55.62 & 51.45 & 36.03 & 30.61 & - & - \\ 
\midrule
PT & & 78.81 & 68.89 & \textbf{73.51} & 71.12 & 65.46 & \textbf{68.17} & 1 & 76.03 & 69.6 & 61.92 & 57.4 & - & -\\
\midrule
RO & & 90.21 & 87.22 & \textbf{88.69} & 87.78 & 86.59 & \textbf{87.18} & 1 & 87.41 & 86.15 & 90.38 & 90.09 & - & - \\
\midrule 
SL & & 74.81 & 52.29 & \textbf{61.55} & 58.56 & 47.2 & \textbf{52.27} & 2 & 36.36 & 27.59 & 36.36 & 25.93 & - & - \\ 
\midrule
TR & & 87.22 & 47.66 & \textbf{61.63} & 81.4 & 45.85 & \textbf{58.66} & 1 & 66.67 & 63.85 & 49.01 & 46.15 & - & - \\
\midrule
\midrule
Total (official) & & 76.22 & 54.27 & \textbf{63.4} & 66.08 & 51.82 & \textbf{58.09} & 1 \\
Total (unofficial) & & 76.86 & 55.58 & \textbf{64.51} & 65.72 & 52.49 & \textbf{58.36} & 1 \\

\bottomrule
\end{tabular}
\begin{tablenotes}
      \small
      \item * At the time of evaluation for the shared task, due to lack of time, we trained the system only for 50 epochs for EN, FR and IT. Here, for these three languages, the results are un-official since we trained the best system (ConvNet+LSTM) with 100 epochs after the evaluation phase. For DE, we trained ConvNet+LSTM+CRF for the evaluation phase. However, after evaluation, we saw that the results of the ConvNet+LSTM system is better also for DE. So, we report the un-official result for DE as well.
    \end{tablenotes}
\end{threeparttable}
\end{table}

In Table \ref{t:results-test}, we report the results of our system (SHOMA) on the blind test data of the Parseme shared task. For each language, we present the general token-based and MWE-based precision, recall and F1-measure and also F1-measures for three selected types of MWEs: LVC.full, VID and VPC.full. Token-based and MWE-based F-measures are indicated with T and M respectively in the right side of the table. More details on the results are published by the shared task organisers.\footnote{\url{http://multiword.sourceforge.net/PHITE.php?sitesig=CONF&page=CONF_04_LAW-MWE-CxG_2018___lb__COLING__rb__&subpage=CONF_50_Shared_task_results}} 
According to the table and the results published for the shared task, on average, our system ranked the best among the participating systems in the open track and also its performance is the best among participating systems in both tracks. 
We report the ranking (\#) of our system in comparison with all participating systems in both tracks rather than only the ones in the open track.

\newcite{Berk-W18} applied a similar neural architecture (Deep-BGT) which is a combination of LSTM and CRF and employs the same pretrained sub-word embedding as ours. The differences are the addition of ConvNet in our system and the labelling format that they applied. Our system outperforms Deep-BGT in 8 out of 10 languages for which they have reported their results. 
It is worth noting that, in contrast to the top systems which participated in the closed track \cite{Waszczuk2018,Stodden2018},
we do not use the dependency parsing tags
that are available for the dataset of most languages in the shared task. This information is not always available in real-word applications when dealing
with running text, and a model that operates independently of this information has its particular advantages. In the case of EL, FR, HE, PT and TR, our system works significantly better than the best system in closed track. 

It is worth noting that we tried the transition-based system for the test data of the five languages (ES, EN, FR, DE, FA) as well. In terms of MWE-based measure our system beats TRANSITION only for the French data out of the five. This shows that the transition-based system would outperform all participating systems. We further demonstrate the effectiveness of our model in Section \ref{sec:generalisation}.

\subsection{Generalisation}
\label{sec:generalisation}

One advantage of our system over a well-performing system like the transition-based is its generalisabilty.   
In order to better understand the robustness of the system, we evaluate the results on seen and unseen expression types separately. We report these results for five languages in Table \ref{t:results-unseen}. The performance of the system on unseen expression types shows its strength in learning across MWE types and its overall generalisation ability. 

\begin{table}[!h]
\centering
\small
\caption{\label{t:results-unseen} Proportion of Seen and Unseen VMWEs in gold standard data and F1 scores of the systems for the two groups individually.}
\setlength{\tabcolsep}{0.2em}
\begin{tabular}{ c c c c c c c c c c c c}
\toprule
 & Seen-Unseen & System & 
 & Seen & Unseen &  \\
 & proportion & & 
 & F1 & F1 & \\
\midrule
 \multirow{2}{*}{ES} &  \multirow{2}{*}{59-41\%} & ConvNet+LSTM & 
 &   \multirow{1}{*}{53.58} &  \multirow{1}{*}{\textbf{19.90}}  \\
 \multirow{2}{*}{} & & Transition  & 
 & 55.24  & 17.77  \\
 \multirow{2}{*}{EN} &   \multirow{2}{*}{29-71\%} & ConvNet+LSTM & 
 &  \multirow{1}{*}{46.70} &  \multirow{1}{*}{\textbf{16.45}}  \\ 
 \multirow{2}{*}{} & & Transition  & 
 & 60.84  & 13.30  \\
 \multirow{2}{*}{FR} &  \multirow{2}{*}{52-48\%} & ConvNet+LSTM & 
 &  \multirow{1}{*}{82.92} &  \multirow{1}{*}{\textbf{36.32}}  \\ 
 \multirow{2}{*}{} & & Transition  & 
 & 84.44  & 16.94  \\
 \multirow{2}{*}{DE} &  \multirow{2}{*}{53-47\%} & ConvNet+LSTM & 
 &  \multirow{1}{*}{72.48} &  \multirow{1}{*}{\textbf{20.00}}  \\ 
 \multirow{2}{*}{} & & Transition  & 
 & 78.88  & 14.84  \\
 \multirow{2}{*}{FA} &  \multirow{2}{*}{66-34\%} & ConvNet+LSTM & 
 &  \multirow{1}{*}{87.73} &  \multirow{1}{*}{\textbf{59.74}}  \\ 
 \multirow{2}{*}{} & & Transition  & 
 & 89.18  & 52.26  \\
\midrule
Avg &    & &  & 69.11 & \textbf{27.49} \\
\bottomrule
\end{tabular}
\end{table}


The performance of our system on unseen
VMWEs is significantly higher than the TRANSITION system and also other participating systems across all target languages. The strength of the transition-based system is on previously seen data. That causes the system to be prone to overfitting. We believe that having the pretrained weights fixed during training is beneficial for the system's robustness. This is also noted by \newcite{howard2018fine} where they gradually update (unfreeze) the initial weights of the network. It is worth noting that the Deep-BGT system \cite{Berk-W18} also shows promising results for un-seen data of the languages for which it is applied to. This suggests the effectiveness of the pre-trained embedding which is common between Deep-BGT and our approach.

\subsection{Effects of Labelling Format}

In Table \ref{t:results-test-filtering}, we report the results based on the shared task evaluation in which there is no distinction among different components of VMWEs. We compare the performance with the results of our proposed IOB-like labelling format where a filtering is applied when the beginning-continuation structure is not satisfied. Our technique considers any singular \texttt{I} as a mistake of the model. The results are reported when applying ConvNet+LSTM approach on the test data. We show the results on a sample of selected languages to analyse the possible effects on model performance.

\begin{table}[h]
\centering
\small
\caption{\label{t:results-test-filtering} The effects of labelling format.}
\setlength{\tabcolsep}{0.2em}
\begin{tabular}{l l c c c c | c c c }
\toprule
\multirow{2}{*}{} & \multirow{2}{*}{Model} &  & \multicolumn{3}{c}{Token-based} & \multicolumn{3}{c}{MWE-based}  \\
  & & & P & R & F1 & P & R & F1 \\
\midrule
\multirow{2}{*}{EN} & with filtering &  & 60.36 & 18.77 & 28.63 & 48.40 & 18.16 & \textbf{26.42}  \\
\multirow{2}{*}{} & no filtering &  & 57.74  & 20.24  & \textbf{29.97}  & 39.57  & 18.16  & 24.90  \\
\midrule
\multirow{2}{*}{FR} & with filtering &  & 82.94 & 57.73 & 68.08 & 72.39 & 54.22 & \textbf{62} \\
\multirow{2}{*}{} & no filtering &  & 80.73 &	 60.12 &	\textbf{68.92} & 63.53 &	54.22 &	58.50 \\
\midrule
\multirow{2}{*}{DE} & with filtering &  & {76.22} & {41.74} & {{53.94}} & {62.38} & {39.96} & {\textbf{48.71}}\\
\multirow{2}{*}{} & no filtering &  & 73.64 &	 44.41 &	\textbf{55.41} & 54.32 &	40.36 &	46.31 \\
\bottomrule
\end{tabular}
\end{table}

According to Table \ref{t:results-test-filtering}, the results computed without filtering non-complete MWEs are slightly higher than the results for our filtering-based approach in terms of token-based F1 score. Performing no filtering increases the recall while adversely affecting precision in all cases. Regarding MWE-based evaluation, the recall scores would not change as is expected, but the considerable drop in precision causes a significant decrease in MWE-based F1 scores.

Based on these results, we believe that this IOB-like labelling scheme is more effective for automatically tagging VMWEs. It is also interesting to apply more complex labelling schemes in future.

\subsection{Our System in the Closed Track}

In order to see the effectiveness of the proposed neural architecture regardless of the pre-trained embedding, we perform a preliminary experiment by using a randomly initialised embedding layer rather than the pre-trained one. We try the model for EN, ES, FR, FA and DE, and the results are presented in Table \ref{t:results-closed}. 

\begin{table}[h]
\centering
\small
\caption{\label{t:results-closed} The performance of our model without pre-trained embeddings.}
\begin{tabular}{l c c c c c c c }
\toprule
 & ES & EN & FR & DE & FA \\ 
\midrule
MWE-based & \textbf{25} & 16.88  & \textbf{53.35} & \textbf{36.30} & 68.20\\
token-based & \textbf{34.74} & \textbf{25.03} & \textbf{63.21} & \textbf{47.78} & 77.66\\
\bottomrule
\end{tabular}
\end{table}

The model outperforms other purely neural models in the shared task \cite{Boros-W18,Zampieri-W18,Ehren-W18}, in the case of 4 out of 5 languages.\footnote{We do not compare with TRAPACC \cite{Stodden2018}, since it is a transition-based system benefiting from dependency-parse information.} In order to apply the architecture to the closed track more investigation on hyper-parameter tuning is required.

\section{Error Analysis}

The ConvNet+LSTM system that we proposed in this paper, on average, achieved the best performance in VMWE identification for different languages. The simple inclusion of a CRF layer (as an activation) to the system did not improve the results. However, we did notice a faster convergence during the training phase. Adding CRF to the network need to be investigated further in future. As for the English data for which our system performs poorly, we take a closer look at the data and we observe that a higher proportion of annotated VMWEs are LVCs and the components of most LVC occurrences have distances of more than one word. The weakness of our system is on cases with bigger gaps and this can explain the low results for English. In general, the system seems to perform inefficiently for discontinuous expressions. For instance, the following LVCs have not been identified using the system. 
\begin{itemize}
\item EN: \textit{\textbf{gave} him a vicious \textbf{stare}}
\item FR: \textit {la \textbf{conduite} des travaux est \textbf{men\'ee}}   
\end{itemize}

\section{Conclusions}
In this paper we proposed a neural model for MWE identification consisting of ConvNet and LSTM layers and an optional CRF layer. We compared the results of the models with strong baselines and the state-of-the-art. Overall, We conclude that a simple addition of a CRF layer to
a well-performing ConvNet+LSTM network does not necessarily improve the results. The ConvNet+LSTM model benefiting from pre-trained embeddings achieved the best performance across multilingual datasets of the Parseme shared task on identification of verbal MWEs. The system performs best in recognising un-seen MWEs which is noteworthy.


\bibliographystyle{acl}
\bibliography{vmwe}

\end{document}